\documentclass{article}

\usepackage{spconf,amsmath,graphicx}
\usepackage[utf8]{inputenc}
\usepackage{amsmath}
\usepackage{algorithm}
\usepackage{algorithmic}
\usepackage{commath}
\usepackage{float}
\usepackage[british]{babel}

\usepackage{cleveref}

\usepackage{enumitem}
\usepackage{xcolor}

\title{\bf GeoSP: A parallel method for a cortical surface parcellation based on geodesic distance}

\name{\begin{tabular}[t]{c@{\extracolsep{1em}}c@{\extracolsep{1em}}c@{\extracolsep{1em}}c}
  \multicolumn{3}{c}{Narciso López-López$^{1,2}$, Andrea Vázquez$^{1}$, Cyril Poupon$^{3}$,}\\
  \multicolumn{3}{c}{Jean-Fran\c{c}ois Mangin$^{3}$, Susana Ladra$^{2}$ and Pamela Guevara$^{1}$}
  \thanks{This work has received funding by the ANID PFCHA/DOCTORADO NACIONAL/2016-21160342, ANID FONDECYT 1190701, ANID PIA/Anillo de Investigaci\'on en Ciencia y Tecnolog\'ia ACT172121, ANID Basal Project FB0008, and by the European Union’s Horizon 2020 research and innovation programme under the Marie Sklodowska-Curie grant agreement No 690941. This work was also partially funded by the Human Brain Project, funded from the European Union’s Horizon 2020 Framework Programme for Research and Innovation under the Specific Grant Agreements No. 785907 (SGA2) and No: 604102 (SGA1).}
\end{tabular}}

\address{$^{1}$ Faculty of Engineering, Universidad de Concepci\'on, Concepci\'on, 4030000, Chile \\
$^{2}$Department of Computer Science, centro de investigaci\'{o}n CITIC, \\ Universidade da Coru\~na, A Coru\~na, 15071, Spain \\
$^{3}$Neurospin, CEA, Paris-Saclay University, Gif-sur-Yvette, 91191, France}

\begin{document}

\maketitle

\begin{abstract}
	
We present \textit{GeoSP}, a parallel method that creates a parcellation of the cortical mesh based on a geodesic distance, in order to consider gyri and sulci topology. The method represents the mesh with a graph and performs a K-means clustering in parallel. It has two modes of use, by default, it performs the geodesic cortical parcellation based on the boundaries of the anatomical parcels provided by the Desikan-Killiany atlas. The other mode performs the complete parcellation of the cortex. Results for both modes and with different values for the total number of sub-parcels show homogeneous sub-parcels. Furthermore, the execution time is 82\,s for the whole cortex mode and 18\,s for the Desikan-Killiany atlas subdivision, for a parcellation into 350 sub-parcels. The proposed method will be available to the community to perform the evaluation of data-driven cortical parcellations. As an example, we compared GeoSP parcellation with Desikan-Killiany and Destrieux atlases in 50 subjects, obtaining more homogeneous parcels for GeoSP and minor differences in structural connectivity reproducibility across subjects.

\end{abstract}

\section{Introduction}
\label{sec:intro}

Magnetic Resonance Imaging (MRI) allows the study of the brain in a non-invasive and in-vivo way. In particular, structural MRI (sMRI) 
gives an anatomical differentiation of main brain tissues, enabling the automatic segmentation of them. The cortical surface can be extracted by softwares like FreeSurfer\footnote{https://surfer.nmr.mgh.harvard.edu/fswiki} \cite{c1,c2} or BrainVISA\footnote{http://brainvisa.info} \cite{c3}. 

A cortex parcellation, i. e. a subdivision of the cortex into several parcels or regions \cite{c4}, can be performed based on different criteria, mostly based on anatomical, functional or diffusion-based information. This is a very complicated task due to the restrictions of each modality and the high inter-subject variability that exists, in particular, in white matter (WM) and gray matter (GM).

When studying the human connectome, brain region definition takes an important role in the study of brain connectivity and function \cite{c5}. 
Anatomical parcellation methods take into account the macroscopical anatomy, like the gyri and sulci \cite{c6, c7}.
For example, Cachia et al. used a geodesic distance to label the cortex mesh vertices, using two nested Vorono\"{i} diagrams and labeled sulci \cite{c7}. Other method uses a statistical surface-based atlas, which includes information of the cortex curvature and the manual labeling of 35 regions of interest (ROIs) per hemisphere \cite{c8}. 

In order to evaluate diffusion-based \cite{c9} or functional-based \cite{c10} parcellations of the cortical surface, these can be compared to a geodesic parcellation which is based on the geodesical properties of the mesh. However, this calculation can be time-consuming.
Therefore, in this work, we propose a parallel method for the complete parcellation of the cortical surface, based on the geodesic distance. The goal is to create a fast individual cortical parcellation, available to the community for parcellation comparisons. The algorithm can be applied to subdivide each anatomical parcel given by Desikan-Killiany (\textit{DK}) atlas, or to  perform the cortical parcellation of the entire brain, depending on the method to be evaluated. 

\section{Materials and Methods}
\label{sec:matmet}

\subsection{Database and tractography datasets} 
\label{ssec:dbtr}

We took from the ARCHI database \cite{c11} 50 subjects for the experiments.
It was acquired with a 3T MRI scanner (Siemens, Erlangen). The MRI protocol used an MPRAGE sequence (160 slices; matrix=256x240; voxel size=1x1x1.1\,mm), including the acquisition of a T1-weighted dataset. 
BrainVISA software was used to pre-process the images. Then, FreeSurfer was applied to calculate the cortical mesh and to obtain the automatical labeling of the cortical regions, according to \textit{DK} atlas. The database also contains deterministic tractography datasets, based on a SS-EPI single-shell HARDI acquisition along 60 optimized DW directions, b=1500 $s/mm^2$ (70 slices, matrix=128x128, voxel size=1.7x1.7x1.7\,mm). 

The experimental procedures involving human subjects described in this paper were approved by the Local Ethical Committee, ``Comité de Protection des Personnes Ile-de-France VII", with codes CPP100002/CPP10002, and all subjects signed an informed consent before inclusion.

\subsection{GeoSP: geodesic cortical surface parcellation}
\label{ssec:app}

The method implemented is called \textit{GeoSP}, and performs the cortical parcellation based on a geodesic distance over the surface. The algorithm has two different modes. The default mode is based on the \textit{DK} atlas to delimit the anatomical parcels and performs a geodesic subdivision of each anatomical parcel. Note that other atlases could also be used. The second mode creates a cortical parcellation for the entire brain. 
For the first mode, the method receives for each anatomical parcel (35 in total) a value $k$, used to divide each anatomical parcel into the specified $k$ sub-parcels (for both hemispheres), i.e. an anatomical parcel with $k = 2$ will be divided into two sub-parcels. On the other hand, the second mode receives a unique $k$ value, which will be used to divide each brain hemisphere into $k$ sub-parcels, based on a geodesic distance, without using any other cortical parcellation.
The method can be subdivided into two main steps: \textit{(1)} a pre-processing that creates a graph representation of the mesh, and \textit{(2)} K-means clustering based on geodesic distance over the mesh \cite{c12}.\\

\begin{figure}[t!]
	\centering
	\includegraphics[scale=.28]{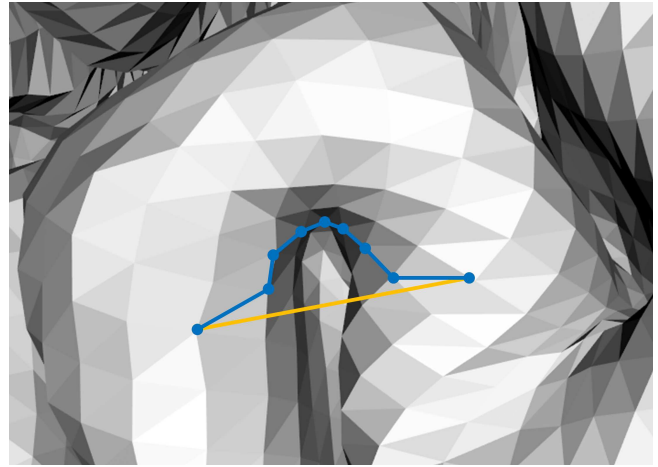}
	\caption{Euclidean distance versus geodesic distance. By using the Euclidean distance between two points, a straight line is obtained (orange path). While the geodesic distance considers the shortest path across the gyri and sulci of the mesh (blue path).}
	\label{fig:edvsgeo}
\end{figure}

\begin{algorithm}[ht]
	\caption{Parallel\_kmeans}
	\label{alg:parallel}
	\begin{algorithmic}[1]
		
		\STATE{$groups \leftarrow [ ]$} \COMMENT{list of sub-parcels containing the indexes of the vertices}
		\STATE{$\textbf{if}$ $k>1$ $\textbf{then}$} \COMMENT{number of clusters}
		
		\STATE\hspace{\algorithmicindent} {$centers \leftarrow initialize()$} \COMMENT{initializing centroids}
		\STATE\hspace{\algorithmicindent} {$centers\_old \leftarrow centers$} 
		\STATE\hspace{\algorithmicindent} {$converge \leftarrow FALSE$}
		\STATE\hspace{\algorithmicindent} {$i \leftarrow 0$}
		
		\STATE\hspace{\algorithmicindent} {$\textbf{while}$ $i$ $\le$ $nIter$ \textbf{and} $!converge$} \COMMENT{iterations and criterion for convergence}
		\STATE\hspace{0.8cm} {$groups \leftarrow calc\_groups()$} \COMMENT{calculating groups}
		\STATE\hspace{0.8cm} {$centers \leftarrow comp\_centroids()$} \COMMENT{computing centroids}
		\STATE\hspace{0.8cm} {$converge \leftarrow stop\_critery()$}
		\STATE\hspace{0.8cm} {$centers\_old \leftarrow centers$}
		\STATE\hspace{0.8cm} {$i \leftarrow i+1$}
		\STATE\hspace{\algorithmicindent} {$\textbf{end while}$}
		
		\STATE{$\textbf{else}$} 
		\STATE\hspace{\algorithmicindent} {$groups \leftarrow [all\_indices]$} \COMMENT{all the indices for one anatomic parcel}
		\STATE {$\textbf{end if}$}
		\RETURN ${groups}$ \COMMENT{returns the list whose elements are the sub-parcel groups(indices))}
		
	\end{algorithmic}
\end{algorithm}

\noindent 
\textbf{1) Pre-processing}\\

Each anatomical parcel (for default mode) or each hemisphere (for the second mode) is represented with an undirected graph. The graph $G = (V,E)$ directly represents the mesh structure, formed by the vertices $V$ and the edges $E$ that join them. 
For the default mode, that performs the subdivision of each anatomical parcel given by the \textit{DK} atlas, the labels of each region are used to identify each parcel.
Finally, Euclidean distance ($d_E$) is calculated between each pair of vertices to create weighted graphs.\\

\noindent 
\textbf{2) K-means clustering}\\
To subdivide an anatomical parcel or a hemisphere into $k$ sub-parcels, a K-means clustering is applied. The algorithm consists of the following sub-steps: \textit{(a)} initializing centroids, \textit{(b)} (re)calculating groups and \textit{(c)} (re)computing centroids. The algorithm uses a parallel implementation and its pseudocode is shown in Algorithm \ref{alg:parallel}. For default mode, the method launches the K-means algorithm in parallel for each anatomical parcel given by \textit{DK} atlas, while for the second mode, it launches a single thread per hemisphere. To exploit the capabilities of parallelism, it is launched in the sub-step of \textit{(re)computing centroids}.

\begin{figure*}[ht!]
	\centering
	\includegraphics[scale=1.06]{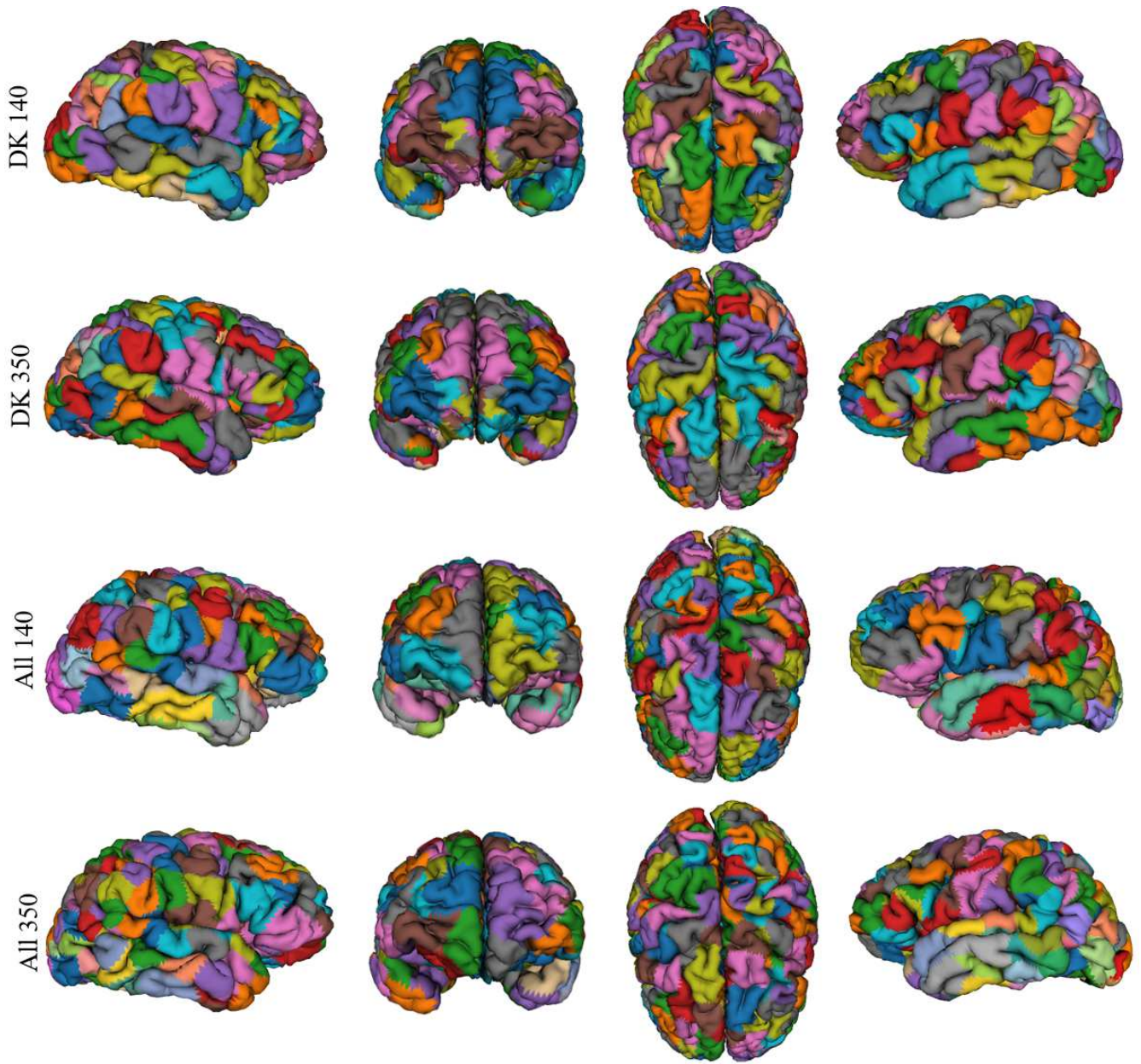}
	\caption{Parcellation of the cortical mesh obtained in one subject for modes based on \textit{DK} atlas and for the entire cortex. Right sagittal, coronal, axial and left sagittal views. First and second rows: parcellation based on \textit{DK} atlas subdivided into 140 (first row) and 350 (second row) sub-parcels, with execution times of 42.9s and 18.1s respectively. Third and fourth rows: parcellation for the entire cortex into 140 (first row) and 350 (second row) sub-parcels, with execution times of 75.4s and 82.25s, respectively.}
	\label{fig:alldesikan}
\end{figure*}

\begin{figure*}[ht!]
	\centering
	\includegraphics[scale=1.04]{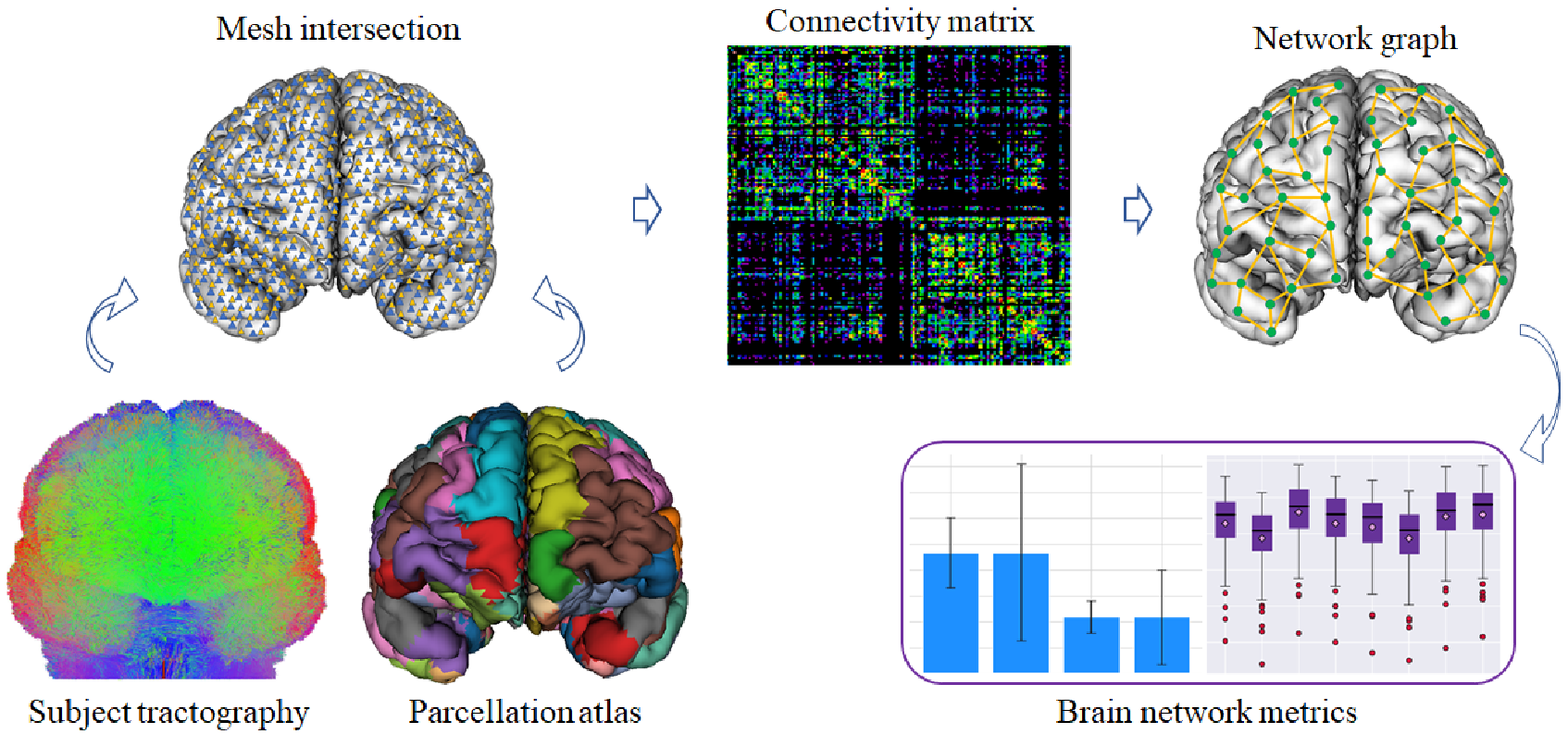}
	\caption{Brain connectivity analysis scheme. First, the tractography of each subject in T1 space is intersected with the cortical mesh, which is parcellated based on an atlas. Then, a connectivity matrix is created containing for each cell, corresponding to a pair of sub-parcels, the total number of connections between them. The matrix is then binarized, indicating with a 1 if the sub-parcels are connected and with a 0 if there is no connectivity between them. The connectivity matrix is finally converted to a network graph to analyze network metrics as the Dice coefficient.}
	\label{fig:graph_cons}
\end{figure*}

\begin{enumerate}[label=(\alph*)]
	\item \textbf{Initializing centroids}. To perform this sub-step,  K-means++ algorithm \cite{c13} is used to select the initial centroids. It has a low time complexity of $\mathcal{O}(log k)$. First, the method receives $k$, the number of sub-parcels (clusters) to divide each anatomical parcel (or each hemisphere). For each anatomical parcel, there may be different $k$. Also, $k$ can be randomly set. Although the selection of starting centroids takes additional time, by using K-means++ the convergence of K-means occurs quickly with reduced computation time. This leads to initial centroids better distributed than random selection across the anatomical parcels.

	\item \textbf{(Re)Calculating groups}. In this sub-step, clusters are (re)calculated by assigning each vertex to the closest centroid. To achieve this, the \textit{single-source shortest path problem} ($SSSP$) is used, which looks for the shortest path from a vertex $v$ (centroid) to the rest of the vertices of the graph $G$. 
	To calculate the distance between vertices, instead of the Euclidean distance, the geodesic distance is used. Then, based on an implementation of the \textit{Dijkstra algorithm with Fibonacci heaps} \cite{c14}, the $SSSP$ is calculated for all the centroids, that is, the shortest path from each centroid to all other vertices. This algorithm runs with low complexity ($\mathcal{O}(|E| + |V| log |V|)$). Finally, for each graph vertex, the distances obtained to the different centroids are compared, and each vertex is assigned to the centroid with the smallest geodesic distance. Figure \ref{fig:edvsgeo} illustrates the Euclidean and geodesic distances for two vertices over the mesh. The path between two points for Euclidean distance is a straight line while the path for the geodesic distance is a route along the surface of the mesh, taking into account the gyri and the sulci.

	\item \textbf{(Re)Computing centroids}. This is the last sub-step of the algorithm, in which the centroids of the clusters must be (re)calculated. First, the \textit{all-pairs shortest paths} problem has to be solved, that is, for each pair of vertices, the shortest path has to be calculated. This is done with the \textit{Floyd–Warshall algorithm} \cite{c15}, which runs in $\mathcal{O}(|V|^3)$. Although the temporal complexity of this step is high, it is still a polynomial running time (cubic) depending on the size of the input. The result obtained is a new centroid, which is the vertex closest to all other vertices in the cluster.	
\end{enumerate}

Sub-steps (b) and (c) are executed until the convergence criterion is reached. For this, the centroids of the current iteration are compared with the previous one. The algorithm stops if the distance is less than 2\,mm or a maximum of 20 iterations is reached.

\section{Results}
\label{sec:res}
The experiments were performed on a computer with an Intel Core i7-8700K 6-core 3.70 GHz CPU, 32 GB of RAM and 12 MB of shared L3 cache. The programming language used is Python 3.6 and the operating system is Ubuntu 18.04.2 LTS with kernel 4.15.0-74. The code is freely accessible at https://github.com/andvazva/GeoSP.

First, Figure \ref{fig:alldesikan} displays the results for one subject with 140 sub-parcels and 350 sub-parcels, for both modes of the method. To obtain 140 sub-parcels using the \textit{DK} atlas, we divide each anatomical parcel into $k = 2$ sub-parcels. Since \textit{DK} atlas has $35$ anatomical parcels per hemisphere, with $k = 2$, we obtain $70$ sub-parcels per hemisphere, leading to a total of 140 sub-parcels for the whole brain. Following the same procedure, to obtain 350 sub-parcels we divide each anatomical parcel into $5$ sub-parcels, which generates $175$ sub-parcels per hemisphere. It can be seen that the method generates homogeneous parcels both for the entire cortex and for the \textit{DK} atlas-based parcellation.

\begin{figure}[ht!]
	\centering
	\includegraphics[scale=1.1]{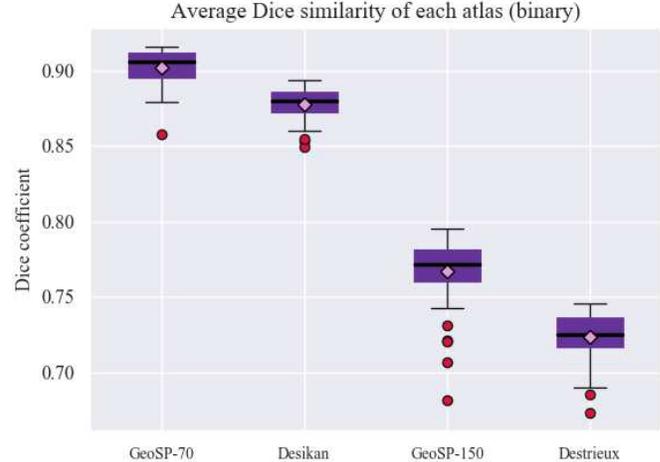}
	\caption{Comparison of the structural connectivity reproducibility between \textit{GeoSP} and the two atlases, with equal number of parcels. X-axis shows the different atlases used. Y-axis contains the Dice coefficient, the closer to one, the greater the reproducibility. The rhombus indicates the mean and the black line the median for each atlas. Results show a difference of $0.024$ between \textit{GeoSP} and \textit{DK} atlas, and $0.043$ between \textit{GeoSP} and \textit{Destrieux} atlas.}
	\label{fig:dice}
\end{figure}

Then, to illustrate an example of use, we calculated the reproducibility of structural connectivity across subjects for three different parcellations: \textit{GeoSP}, \textit{DK} and \textit{Destrieux}.
Figure \ref{fig:graph_cons} displays the scheme of processing performed to obtain the reproducibility analysis for a parcellation. For each subject, we used the tractography dataset in T1 space to calculate the structural connectivity matrix, based on each parcellation. To construct a matrix, the intersection of the fibers with the cortical mesh is determined and the labels of the pair of parcels connected by each fiber are used to add a count in the corresponding cell of the matrix. Next, the matrx is binarized and converted into a graph to use network metrics. One of these metrics is the Dice coefficient and was calculated between each pair of subjects, for each method. Figure \ref{fig:dice} shows a boxplot of the reproducibility among the 50 subjects between \textit{GeoSP} and the other anatomical atlases. The reproducibility is slightly higher for \textit{GeoSP} in both cases, with a difference of $0.024$ between \textit{GeoSP} and \textit{DK} (70 parcels) and of $0.043$ for \textit{GeoSP} and \textit{Destrieux} (150 parcels).

\begin{figure}[ht!]
	\centering
	\includegraphics[scale=1.1]{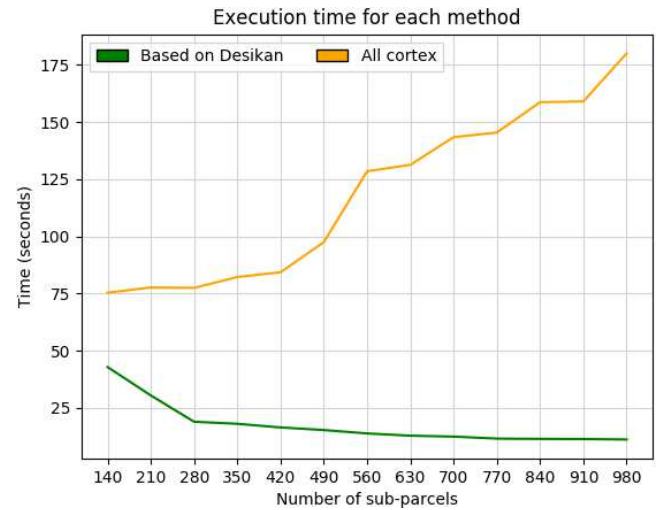}
	\caption{Execution time (seconds) for each mode, depending on the number of sub-parcels. As expected, the subdivision into sub-parcels according to the delimitation given by the Desikan-Killiany atlas is less expensive than subdividing the entire cortex.}
	\label{fig:times}
\end{figure}

Finally, the execution time for both modes was compared. Figure \ref{fig:times} displays the execution times according to the number of sub-parcels in which the cortex is subdivided. For mode one, based on \textit{DK} parcellation, the execution time decreases with the number of parcels. This is because the greater the number of sub-parcels, and being delimited by the anatomical parcels of the atlas, the algorithm has to perform fewer computations when recomputing the centroids. On the other hand, for the entire cortex, with a greater number of sub-parcels, more time is needed to subdivide the cortex. This is due to the size of the graphs (one for each hemisphere), where the recalculation of centroids becomes very expensive since it is necessary to recalculate all the shortest paths between all the pairs.

\addtolength{\textheight}{-2.5cm}

\section{Conclusions}
\label{sec:con}

We propose a parallel method to perform a parcellation of the cortical surface mesh based on geodesic distance. 
The algorithm was tested in 50 subjects. Results show homogeneous sub-parcels for both modes and different number of sub-parcels.
Structural connectivity reproducibility between \textit{GeoSP} and two anatomical atlases is very similar and slightly higher for \textit{GeoSP}. This may be due to the higher homogeneity of the parcels with \textit{GeoSP}. Moreover, the greater the number of parcels, the less reproducibility will be obtained. Hence, this test shows that special attention should be given to the indices to be used in comparisons between parcellations. In any case, we provide a fast and configurable parcellation method based on geodesic distance, available to the community, to perform the comparison and evaluation of data-driven parcellations, like those based on diffusion or functional MRI.



\begin{thebibliography}{99}
	
	\bibitem{c1} B. Fischl, M. I. Sereno, and A. M. Dale, ``Cortical Surface-Based Analysis,'' NeuroImage, vol. 9, no. 2, pp. 195–207, feb 1999.
	\bibitem{c2} K. Gopinath, C. Desrosiers, and H. Lombaert,
	``Graph convolutions on spectral embeddings for cortical surface parcellation,'' Medical Image Analysis, vol. 54, pp. 297–
	305, 2019.
	\bibitem{c3} L. Brun, A. Pron, J. Sein, C. Deruelle, and O. Coulon, ``Diffusion MRI: Assessment of the Impact of Acquisition and Preprocessing Methods Using the	BrainVISA-Diffuse Toolbox,'' Frontiers in Neuroscience, vol. 13, pp. 536, 2019.
	\bibitem{c4} F. Zhang, Y. Wu, I. Norton, Y. Rathi, A. J. Golby, and L. J. O’Donnell, ``Test–retest reproducibility of white matter parcellation using diffusion MRI tractography fiber clustering,'' Human Brain Mapping, vol. 40, no. 10, pp. 3041–3057, 2019.
	\bibitem{c5} O. Sporns, G. Tononi, and R. Kotter, ``The human connectome:	a structural description of the human brain,'' PLOS Computational Biology, vol. 1, no. 4, pp. e42, 2005.
	\bibitem{c6} C. Destrieux, B. Fischl, A. Dale, and E. Halgren, ``Automatic parcellation of human cortical gyri and sulci using standard
	anatomical nomenclature,'' NeuroImage, vol. 53, no. 1, pp. 1–
	15, 2010.
	\bibitem{c7} A. Cachia, J.-F. Mangin, D. Rivière, D. Papadopoulos-Orfanos, F. Kherif, I. Bloch, and J. Regis, ``A generic framework for the parcellation of the cortical surface into gyri using geodesic Voronoï diagrams,'' Medical Image Analysis, vol. 7, no. 4, pp.
	403–416, 2003.
	\bibitem{c8} R. S. Desikan, F. Ségonne, B. Fischl, B. T. Quinn, B. C. Dickerson, D. Blacker, R. L. Buckner, A. M. Dale, R. P. Maguire, B. T. Hyman, et al., ``An automated labeling system for subdividing the human cerebral cortex on MRI scans into gyral based regions of interest,'' NeuroImage, vol. 31, no. 3, pp. 968–980, 2006.
	\bibitem{c9} S. N. Sotiropoulos and A. Zalesky, ``Building connectomes using diffusion MRI: why, how and but,'' NMR in Biomedicine, vol. 32, no. 4, pp. e3752, 2019.
	\bibitem{c10} K. Dadi, M. Rahim, A. Abraham, D.	Chyzhyk, M. Milham, B. Thirion, G. Varoquaux, Alzheimer’s Disease Neuroimaging Initiative, et al., ``Benchmarking functional connectome-based predictive models for
	resting-state fMRI,'' NeuroImage, vol. 192, pp. 115–134, 2019.
	\bibitem{c11} B. Schmitt, A. Lebois, D. Duclap, P. Guevara, F. Poupon,
	D. Rivière, Y. Cointepas, D. LeBihan, J.-F. Mangin, and C. Poupon, ``CONNECT/ARCHI: an open database to infer atlases of the human brain connectivity,'' in ESMRMB, 2012.
	\bibitem{c12} P. Fr{\"a}nti, and S. Sieranoja, ``How much can k-means be improved by using better initialization and repeats?,'' Pattern Recognition, vol. 93, pp. 95–112, 2019.
	\bibitem{c13} S. Lattanzi and C. Sohler, ``A better k-means++ Algorithm via Local Search,'' in Int. Conf. on Machine Learning, 2019, pp. 3662–3671.
	\bibitem{c14} R. A. Chowdhury and V. Ramachandran, ``Cache-Oblivious Buffer Heap and Cache-Efficient Computation of Shortest Paths in Graphs,'' ACM Trans. on Algorithms (TALG),
	vol. 14, no. 1, pp. 1–33, 2018.
	\bibitem{c15} Z. Ramadhan, A. Putera Utama Siahaan, and M. Mesran,
	``Prim and Floyd-Warshall Comparative Algorithms in Shortest Path Problem,'' in Proc. of the Joint Workshop KO2PI and The 1st Int. Conf. on Advance \& Scientific Innovation, 2018.
	
	
\end{thebibliography}

\end{document}